# Facial Soft Biometrics for Recognition in the Wild: Recent Works, Annotation, and COTS Evaluation

Ester Gonzalez-Sosa, Julian Fierrez, *Member, IEEE*, Ruben Vera-Rodriguez, and Fernando Alonso-Fernandez

*Abstract*—The role of soft biometrics to enhance person recognition systems in unconstrained scenarios has not been extensively studied. Here, we explore the utility of the following modalities: gender, ethnicity, age, glasses, beard, and moustache. We consider two assumptions: 1) manual estimation of soft biometrics and 2) automatic estimation from two commercial off-the-shelf systems (COTS). All experiments are reported using the labeled faces in the wild (LFW) database. First, we study the discrimination capabilities of soft biometrics standalone. Then, experiments are carried out fusing soft biometrics with two state-of-the-art face recognition systems based on deep learning. We observe that soft biometrics is a valuable complement to the face modality in unconstrained scenarios, with relative improvements up to 40%/15% in the verification performance when using manual/automatic soft biometrics estimation. Results are reproducible as we make public our manual annotations and COTS outputs of soft biometrics over LFW, as well as the face recognition scores.

## I. INTRODUCTION

SOFT biometrics refer to physical and behavioral traits that can be semantically described by humans [9], [19]. Although soft biometric traits may not possess sufficient distinctiveness or uniqueness to allow highly accurate recognition [4], they can be useful to enhance person recognition under certain conditions. For instance, Dantcheva *et al.* [10] suggested that it might be possible to perform recognition when considering a sufficient number of them, in a *bag of soft biometrics*. In application domains such as surveillance scenarios in which primary biometric traits may suffer from different types of degradation, the use of soft biometrics has been shown to improve the performance attained with primary biometrics (also known as hard biometrics) [27], [31], [38].

Although some works have explored fusion of soft and hard biometrics information [20], [22], [32], [39], [41], there are still some unsolved questions that this work will address, specifically: $i$) which are the most discriminative soft biometric modalities?; $ii$) how accurately can soft biometric modalities be extracted from unconstrained face images in an automatic manner?; and $iii$) to what extent they can improve accuracy of face verification systems in unconstrained scenarios? In turn, we will analyze the usefulness and convenience of soft biometric modalities in unconstrained scenarios, and their fusion with state-of-the-art face verification systems.

In this work we study a set of 6 soft biometric traits, generated both manually and automatically from two Commercial Off-The-Shelf systems (COTS). In order to extract conclusions that can be extrapolated to real situations, we report experiments using the Labeled Faces in the Wild (LFW) database, which is an unconstrained database under challenging conditions such as pose, illumination, expression or occlusion, among others. Two state-of-the art face verification systems based on deep learning are considered in this study, playing the role of hard biometric systems. We highlight the following contributions:

- Comprehensive review of recents works on biometric recognition using soft biometrics extracted from facial images, with special focus in scenarios in-the-wild.
- Empirical assessment of the discrimination capabilities of the following soft biometrics in scenarios in-the-wild: gender, age, ethnicity, glasses, beard and moustache. The assessment compares soft biometrics computed on the popular LFW database with two cost-off-the-shelf systems, in comparison with a generated groundtruth that has been made publicly available.[1]
- Experimental evidence of the improvement of unconstrained facial recognition systems when they are fused with soft-biometric modalities.

This paper is structured as follows. Related works regarding the use of soft biometrics for recognition purposes is covered in Section II. The LFW database and the standard experimental protocol used in this paper is described in Section III.

Manuscript received September 7, 2017; revised December 29, 2017 and February 9, 2018; accepted February 13, 2018. Date of publication February 19, 2018; date of current version April 4, 2018. This work was supported in part by the Spanish Guardia Civil and the project CogniMetrics from MINECO/FEDER under Grant TEC2015-70627-R and in part by the Imperial College London under Grant PRX16/00580. The work of E. Gonzalez-Sosa was supported by a Ph.D. Scholarship from the Universidad Autonoma de Madrid. The work of F. Alonso-Fernandez was supported in part by the Swedish Research Council, in part by the CAISR program, and in part by the SIDUS-AIR project of the Swedish Knowledge Foundation. The associate editor coordinating the review of this manuscript and approving it for publication was Prof. Clinton Fookes. *(Corresponding author: Ester Gonzalez-Sosa.)*

E. Gonzalez-Sosa is with Nokia Bell-Labs, 28045 Madrid, Spain, and also with Escuela Politecnica Superior, Universidad Autonoma de Madrid, 28049 Madrid, Spain (e-mail: ester.gonzalez@nokia-belllabs.com).

J. Fierrez and R. Vera-Rodriguez are with Escuela Politecnica Superior, Universidad Autonoma de Madrid, 28049 Madrid, Spain (e-mail: julian.fierrez@uam.es; ruben.vera@uam.es).

F. Alonso-Fernandez is with School of Information Technology, Halmstad University, 30118 Halmstad, Sweden (e-mail: feralo@hh.se).

Color versions of one or more of the figures in this paper are available online at http://ieeexplore.ieee.org.

Digital Object Identifier 10.1109/TIFS.2018.2807791

[1]See https://atvs.ii.uam.es/atvs/LFW_SoftBiometrics.html

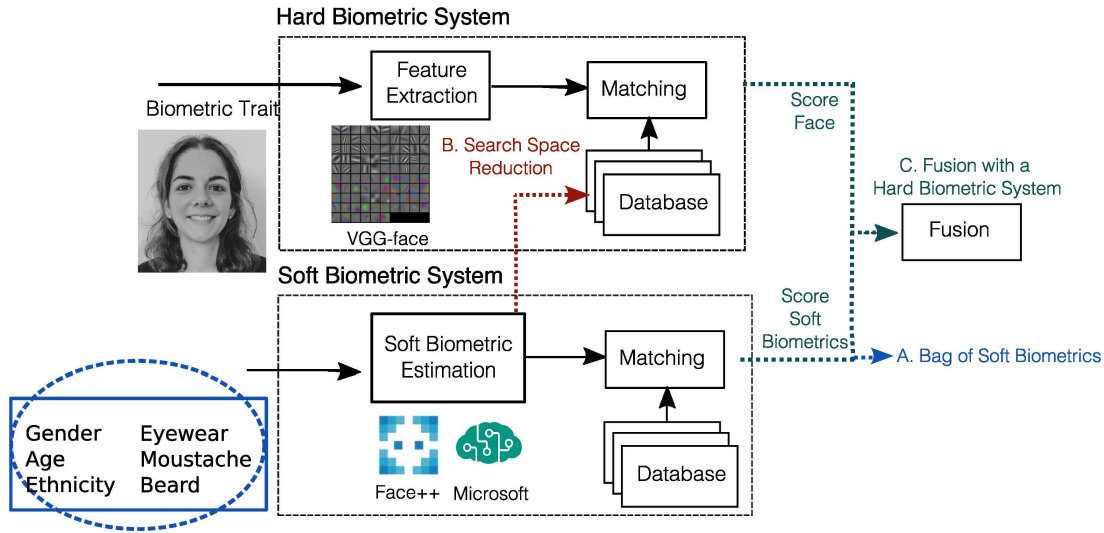

Fig. 1. **Soft Biometrics for Recognition.** Main applications of soft biometrics for recognition: A) *Bag of Soft Biometrics:* using a set (bag) of soft biometrics for recognition in an exclusive manner; B) *Search Space Reduction:* reducing the number of comparisons made by the classifier from the hard biometric system by restricting to those with a certain soft biometric profile, and C) *Fusion with a Hard Biometric System:* fusing soft and hard biometric information.

TABLE I

RELATED WORKS THAT USE A BAG OF SOFT BIOMETRICS TO PERFORM PERSON RECOGNITION ABBREVIATIONS USED: ACCURACY (ACC.); RANK-1 (R-1); EQUALL ERROR RATE (EER); CATEGORIES (CAT.); DEEP CONVOLUTIONAL NEURAL NETWORKS (DCNN). RELATED WORKS ARE PRESENTED CHRONOLOGICALLY

| Work | Features | Manual / Automatic | Dataset | Performance (%) |
|---|---|---|---|---|
| Kumar et al. 2011 [22] | 73 attributes (binary) | Automatic (SVM) | LFW (5749 subjects) | 85.5 (Acc.) |
| Klare et al. 2014 [20] | 46 Forensic facial attributes (categorical) | Test: Manual (Sketches) Train: Automatic (OpenBR) | CUFSF (1194 subjects) | 92.0 (R-1) |
| Reid et al. 2014 [32] | 19 labels from face and body (comparative) | Manual | Multi-Biometric Tunnel (100 subjects) | 95.0 (Acc.) |
| Tome et al. 2015 [39] | 32 (continous) + 24 (categorical) facial soft biometrics | Manual | ATVS Forensic (50 subjects) MORPH (130 subjects) | 3.0 (EER) 12.2 (EER) |
| Arigbabu et al. 2015 [7] | Soft face: facial shape & skin color (continuous) Soft body: Height, & weight (continuous) | Automatic | UPM Soft Bio. (70 subjects) | 88.0 (R-1) |
| Samangouei et al. 2016 [34] | 40 facial attributes (binary) | Automatic (DCNN) | MOBIO (152 subjects) AA01 (50 subjects) | 17.0 (EER) 20.0(EER) |
| Almudhahka et al. 2016 [5] | 24 labels from face and body (comparative) | Manual | LFW (5749 subjects) | 93.6 (Acc.) |
| Ghalleb et al. 2016 [13] | Soft face: skin and hair color (categorical) Soft body: Height, 4 body meas. & gait | Automatic | Face94 & Casia Gait (20 subjects) | 95.0 (R-1) |
| Rudd et al. 2016 [33] | 40 facial attributes (categorical) | Automatic (SVM) | LFW (5749 subjects) | 85.0 (Acc.) |

Section IV introduces the set of soft biometrics analyzed in this work and the COTS employed for their automatic estimation. Section V describes the soft biometric and hard biometric verification systems. The different experiments carried out with soft biometrics, either as a bag of soft biometrics or in combination with hard biometric systems are covered in Section VI. Finally, Section VII concludes the paper with a brief summary and discussion.

## II. RELATED WORKS

In what follows we focus on related works studying the use of soft biometrics for recognition. Related works on automatic algorithms to predict soft biometrics are beyond the scope of this Section. We refer the reader to one of many extensive surveys on the topic, for example [9], [15], [27], [36].

Fig. 1 shows how soft biometrics may help recognition in different ways: *A)* performing recognition using a bag of soft biometric modalities, *B)* reducing the search space of a hard biometric modality, and *C)* complementing the evidence given by hard biometric traits.

### A. Bag of Soft Biometrics

Some previous works have attempted to perform recognition based on a bag of soft biometrics (see Table I). In [22], a set of 73 attribute-based classifiers were used reporting accuracy rates of 85.54% in the LFW database. All attributes were automatically extracted using Support Vector Machines as classifier.

Rudd et al. [33] proposed a joint deep convolutional neural network to estimate 40 facial attributes, addressing also the problem of imbalanced data. The accuracy reported for person recognition on LFW dataset was 85.05%, very similar to the work by [22], but using a lower number of facial attributes (40 as opposed to 73).

The majority of works mentioned in this section have employed facial soft-biometric measurements only. Together with [32], the works [7] and [13] also employed body soft-biometric information for recognition, reporting reasonable rank-1 rates of 88% (using a fuzzy logic matcher) and 95%, respectively.

Soft biometrics have been also studied in forensic contexts. In [20], a framework was proposed to match forensic facial attributes with attributes extracted from hand-drawn police sketches. Later, Tome et al. [39] proposed a set of 32 continuous and 24 discrete facial soft biometric measurements based on geometric information from eyebrows, nose, mouth or eyes, inspired by practical protocols from some international forensic laboratories. Verification results with 3.06% and 12.27% EER were reported on the ATVS Forensic and MORPH databases, respectively. The works by Niinuma et al. and by Samangouei et al. explored the use of a set of facial attributes for active authentication purposes [26], [34] and estimated through deep convolutional neural networks. Then, Reid et al. [32] showed how comparative categorical attributes could be used for identification as a better alternative to absolute categorical attributes employed by previous studies. To this end, they invited volunteers (through crowdsourcing annotation) to manually derive 19 relative measures of facial appearance and body proportions. A recognition accuracy of 95% was attained using the Multi-biometric Tunnel Database. Later, in [5] they enlarged the number of comparative attributes up to 24, by including information regarding geometric measures from the eyebrows reporting an accuracy of 93.66% over the LFW database. Unfortunately, although comparative soft biometric attributes have been shown to provide better accuracy than absolute attributes, automatic extraction of comparative attributes in practical scenarios may be an issue.

### B. Search Space Reduction

In surveillance and forensic scenarios, suspect identification could imply comparing a query CCTV image against a considerable number of subjects in a watch-list database. With the inclusion of soft biometrics, the number of comparisons could be reduced to those subjects with a similar soft biometric profile to the query [10], [41].

One example was presented in the context of the Boston Marathon attacks [21]. In order to prove how face recognition algorithms could have assisted law enforcement in identifying suspects, they build an artificial database where suspect images were injected. Demographic filtering substantially improves retrieval rankings compared to the blind search (e.g. from 12.446 to 1746 for probe image 1a using NeoFace 3.1), with an improvement generally proportional to the reduction in gallery size (from 1.000.000 to less than 170.000)

Givens et al. [14] brought to the fore that certain hard biometric modalities are more difficult to use in some demographic groups, for instance, the hypothetical case that Asian people are more difficult to distinguish based on their fingerprint than Caucasian people. Another example can be iris occlusion due to eyelids, which is more prominent in Asian populations. This can be used to tune some parameters of the hard biometric system based on soft biometric measurements to accomodate to the particularities of certain populations.

Although reducing the search space may reduce the time employed for identification, it is crucial that soft biometric modalities are estimated accurately, as failure detection would imply that genuine suspects are left out of the search space. For example, in the work conducted by Zhang et al. [41] it was observed that the verification rate at $FAR = 0.001$ obtained while pruning with gender and ethnicity information improved from 15.2% to 20.1% when those soft biometrics were manually estimated. However, when automatically estimated, the verification rate worsened down to 12.6% (automatic gender and ethnicity classification results were of 84.3% and 84.2%, respectively).

### C. Fusion of Soft and Hard Biometric Traits

Table II describes previous works that have explored hard and soft biometrics jointly for recognition purposes. Zewail et al. [40] proposed to use iris color as soft information to improve a multimodal biometric system composed of fingerprint and iris texture information. The fingerprint-based system was based on steerable pyramid filters and the iris-based system was based on multi-channel log-Gabor filters. They reported experiments using a chimeric database using the DSP_AAST.vl.0 for irises and the Fingerprint Verification Competition FVC 2002 for fingerprints. Iris color was automatically extracted through color histograms. The reported results were of 69%, 85%, and 99% of Genuine Acceptance Rate for iris, fingerprint and the fusion of iris, fingerprint and iris color, respectively.

In [18] the combination of soft information with face and fingerprint-based biometric systems was assessed. Concretely, LDA-based face and minutiae-based fingerprint systems were combined and tested on a chimeric database with 263 subjects. They integrated primary biometrics with gender, ethnicity and height. Gender and ethnicity were automatically extracted based on a LDA-based algorithm with accuracy rates of 96.3% and 89.6%, respectively; height was labeled manually. Conclusions from this work suggested that $i$) height was more discriminative than gender or ethnicity, and $ii$) fingerprint improved when considering gender and ethnicity but face biometrics only improved with height (they argued that face, gender and ethnicity are not complementary as they are derived from the same facial features). Regarding statement $i$), we believe the fact that height was a continuous variable in comparison with the discrete nature of ethnicity and gender played an important role in their superior discrimination. Even if it is logical that there are only two different values for gender, this work was built using ethnicity information that only reflects if someone was Asian or non Asian. We devise even more improvement over the primary biometric system by being more precise regarding the particular ethnicity. Later, in [3] it was showed how medical soft biometrics such as body weight or fat information automatically estimated may help to improve the performance of a fingerprint commercial system such as Biometrika.

TABLE II
**Related Works That Combine Soft and Hard Biometric Modalities** Abbreviations Used: Accuracy (Acc.); Genuine Acceptance Rate (GAR); Rank-1 (R-1); Total Error Rate (TER); Categories (Cat.); Continuous (Cont.); Histograms (Hist.); Local Region Principal Component Analysis (LRPCA); Linear Discriminant Analysis (LDA), Principal Component Analysis (PCA); Support Vector Machines (SVM); Mean Standard Deviation (MSD). We Indicate Wether Soft Biometrics Are Manually or Automatically Estimated, Including the Accuracy Rate When Reported in the Related Work. Related Works Are Presented Chronologically

| Work | Hard Biometric Modalities | Soft Biometric Modalities | Dataset | Performance (%) |
|---|---|---|---|---|
| Zewail et al. 2004 [40] | Hard1: Fingerprint (steerable pyramid filter) Hard2: Iris texture (Multi-chanell log-Gabor Filters) | Iris color (Automatic, cont.) | Iris: DSP_AAST.v1.0 (21 subjects) Fingerprint: Subset from FVC 2002 (<110 subjects) | GAR for: Hard1: 69.0 Hard2: 85.0 Hard1+ Hard2 + Soft: 99.0 |
| Jain et al. 2004 [18] | Fingerprint (Minutiae-based) | Gender (96.3%. Acc., 2 cat) Ethnicity (89.6% Acc., 2 cat) Height (Manual, cont.) | Chimeric (160 subjects) | R-1 for: Hard: 86.5 Hard + Soft: 90.5 |
| Jain et al. 2004 [18] | Face (LDA) | Height (Manual, cont.) | Chimeric (160 subjects) | R-1 for: Hard: 60.0 Hard + Soft: 65.0 |
| Ailisto et al. 2006 [3] | Fingerprint (Biometrika FX 2000 commercial system) | Body Weight (Automatic, cont.) Fat (Automatic, cont.) | Private (62 subjects) | TER for: Hard: 3.9 Hard + Soft: 1.5 |
| Park et al. 2010 [28] | Face (FaceVACS commercial system) | Gender (Manual, 2 cat) Ethnicity (Manual, 2 cat) Facial Marks (Automatic, cont.) | FERET (1199 subjects) | R-1 for: Hard: 90.6 Hard + Soft: 92.0 |
| Abreu et al. 2011 [1] | Face (PCA + SVM) | Age (10.2% Mean Error, 3 cat.) Gender (9.3% Mean Error, 2 cat.) | BioSecure (600 subjects) | Error mean for: Hard: 8.1 Hard + Soft: 2.3 |
| Abreu et al. 2011 [1] | Fingerprint (VeriFinger Minutiae) | Age (9.3% Mean Error, 3 cat.) Gender (12.8% Mean Error, 2 cat.) | BioSecure (600 subjects) | Error mean for: Hard: 10.8 Hard + Soft: 3.2 |
| Tome et al. 2014 [38] | Face (Sparse Representation) | Age (Manual, 7 cat) Ethnicity (Manual, 7 cat) Gender (Manual, 2 cat) 20 measures from face and body. (Manual) | Multi-Biometric Tunnel Database (50 subjects) | EER for: Hard: 15.96 Hard + Soft: 7.68 |
| Prakash et al. 2014 [30] | Face (Gabor + LBP) | Skin Color (Automatic, Hist.) Clothes (Automatic, Hist.) | Indian Database (61 subjects) | Recognition Rate for: Hard: $N/A$ Hard + Soft: 89.13 |
| Zhang et al. 2015 [41] | Face (LRPRCA & Cohort LDA) | Gender (84.3% Acc., 2 cat) Ethnicity (84.3% Acc., 2 cat) Eye Color (68.0% Acc., 2 cat) Hair Color (76.6% Acc., 2 cat) Eyebrow (66.1% Acc., 2 cat) | Ugly partition from GBU dataset (874 subjects) | VR at $FAR = 0.001$: Hard: 15.2 Hard + Soft: 15.5 Hard + Gender= 16.2 |
| Ghalleb et al. 2016 [12] | Face (Discrete Wavelet Transform) | Skin Color (Automatic, cont.) Body Soft (Side View): Height (Automatic, cont.) Waist Width (Automatic, cont.) Body Soft (Frontal View) Waist length (Automatic, cont.) Leg circ. (Automatic, cont.) Arm circ. (Automatic, cont.) Head width (Automatic, cont.) Shoulder length (Automatic, cont.) | CASIA Gait DatasetA (18 subjects) | EER for: Frontal View: Hard: 19.3 Hard + Soft: 3.0 Side View Hard: 28.0 Hard + Soft: 7.5 |
| Ours | Hard1: Face (Face++) Hard2: Face (VGG-face + cosine) | Gender (Manual, 2 cat.) Ethnicity (Manual, 5 cat.) Age (Manual, 5 cat.) Glasses (Manual, 2 cat.) Moustache (Manual, 2 cat.) | LFW (5749 subjects) | EER for: Hard1: 12.7 Hard1 + Soft: 7.6 Hard2: 7.8 Hard2 + Soft: 4.4 |
| Ours | Hard1: Face (Face++) Hard2: Face (VGG-face + cosine) | Gender (92.9% Acc., 2 cat.) Ethnicity (87.4% Acc., 3 cat.) Age (3.5% MSD, cont.) Glasses (92.2%, 2 cat.) Moustache (94.1% Acc., 2 cat.) | LFW (5749 subjects) | EER for: Hard1: 12.7 Hard1 + Soft: 11.4 Hard2: 7.8 Hard2 + Soft: 6.6 |

In a subsequent work, face information extracted from FaceVACS system was combined with soft biometrics including gender, ethnicity and facial marks (e.g. tattoos) [28]. The authors reported improvement of rank-1 identification results from 90.61% to 92.02% when considering soft biometrics. The authors stated also that facial marks can play an important role when identifying twins. It is worth noting that facial marks were estimated automatically. Later, the work by Abreu et al. [1] showed consistently how age and gender information, automatically estimated, outperformed either a face verification system based on PCA and SVM or a fingerprint verification system based on the VeriFinger commercial system to a great extent. The work by Tome et al. [38] empirically proved the convenience of fusing hard and soft information in scenarios at a distance. Concretely, a manual set of facial and body-based soft biometrics was proposed to complement a sparse representation-based face verification system. They studied different distance scenarios and reported improvements from 15.96% to 7.68% of EER in the most challenging distance-scenario.

Prakash and Mukesh [30] proposed the joint use of face information (using Gabor filters and Local Binary Patterns) with soft information such as clothes and facial skin for continuous user authentication purposes as a two-stage process. First they authenticate the user through face recognition. In case of any variations in soft biometrics, the system checks for hard biometric verification. They reported average results of 89.13% of recognition rate when using one single frontal image for the gallery from Indian Database (containing expression and pose variations).

Zhang et al. [41] explored in more depth the convenience of integrating a set of 5 soft biometrics (gender, ethnicity, eye color, hair color and eyebrow) with hard biometric systems under two assumptions: $i$) perfect knowledge and $ii$) automatic estimations. They reported verification rates of 15.2% at FAR = 0.001, when using a face system composed of the fusion of Local Region Principal Component Analysis (LRPCA) and Cohort Linear Discriminant Analysis (CohortLDA) for the Ugly partition, which is the most challenging partition of the GBU dataset. This verification rate is improved up to 16.4% when introducing gender information or up to 15.5% when introducing the whole set of 5 soft biometrics. However, results were reported using the GBU dataset, Local Region Principal Component Analysis (LRPCA) and Cohort Linear Discriminant Analysis (CohortLDA), which do no represent real unconstrained scenarios nor state of the art techniques.

The work by Ghalleb and Amara [12] explored the use of some facial and body soft information to enhance a face verification system in remote acquisition scenarios. Concretely, the face verification system was based on Discrete Wavelet Transformation. The considered soft biometrics were skin color as facial soft information and some body measurements: height and waist width for side view, and waist length, shoulder width, head width, leg and arm circumferences for frontal view. They reported results using the CASIA Gait Dataset A using 360 images from 18 different people. Equal error rates were reduced from 19.30% to 3.00% and from 28.00% to 7.48% when incorporating soft information to the hard biometric system for the frontal and side view, respectively.

Based on the aforementioned discussion, we concluded that soft biometrics may lead to significant improvements to hard biometrics. However, we believe there are some important issues which have not yet been addressed extensively. In this regard, we here aim to gain insight regarding: $i$) the use of soft biometrics in unconstrained scenarios such as the LFW framework, $ii$) the utility of soft biometrics when combined with hard biometrics from state-of-the-art such as face recognition based on deep learning, and $iii$) the impact of automatic estimation of soft biometrics over person recognition and over fusion with a face recognition system.

## III. THE LFW DATABASE AND EXPERIMENTAL PROTOCOL

Labeled Faces in the Wild is a database of face images designed to study the problem of unconstrained face recognition [24]. It contains 13233 JPEG images of $250 \times 150$ pixels from 5749 different individuals. Images from LFW show a large range of the variation seen in everyday life, including pose, lighting, expression, background, ethnicity, age, gender, clothing, hairstyles, camera quality, color saturation, focus, and other parameters. We decided to use this difficult database to investigate the potential of soft biometrics in unconstrained scenarios. The majority of previous works have used more controlled datasets such as FERET, CASIA, or BioSecure to prove the role of soft biometrics for person recognition.

We report results using the standard experimental protocol from view 2, composed of 10 folds, each of it with 300 genuine and 300 impostor comparisons. Verification results in this database are given using 10-fold cross validation and in terms of Equal Error Rate (EER).

## IV. SOFT BIOMETRICS

In this section we describe the soft biometrics chosen and the manual and automatic procedures to estimate them.

### A. Manual Estimation

Prior to this work there was no groundtruth of soft biometric information from the LFW dataset publicly available to study their impact over recognition. Although it is true that the work by Kumar et al. [22] provides the automatic estimation from their 73 attribute classifiers, they do not make public the corresponding groundtruth. Likewise, the attributes generated by [25] are not available at the time of this current work. Recently Jain et al. [15] have made public demographic annotations from LFW (age, gender, ethnicity), but those annotations lack information regarding facial attributes. As one of the goals of this work is to understand to what extent soft biometrics complement hard biometrics when using manual or automatic estimations, we decided to create our own groundtruth composed of soft biometrics and facial attributes. The groundtruth annotation was carried out by a single person. Table III describes the soft biometrics considered along with their corresponding instances (different values that a soft biometric may have, that have been chosen similar as the work in [23]). Note also that the quantified value of each instance is included in parentheses. We have collected information regarding gender, age, ethnicity, glasses, beard and moustache. In order to reduce as much as possible the subjectivity associated to a single annotator, some annotation criteria were fixed as a reference guide. For instance, we decided that a person was *Caucasian* if the skin color was white; *Black* when then skin color was black; *Asian* when the eyes were slanted; *Indian* for clearly visible Indian face features, and

TABLE III
SOFT BIOMETRICS EXTRACTED FROM THE LFW DATABASE. SOFT BIOMETRICS ARE DESCRIBED THROUGH ALL THEIR POSSIBLE VALUES (INSTANCES) AND CORRESPONDING QUANTIZED VALUES

| **Gender** | Male (0) | Female (1) | | | |
|---|---|---|---|---|---|
| **Age** | Baby (0) | Child (1) | Youth (2) | Middle Aged (3) | Senior (4) |
| **Ethnicity** | Caucasian (0) | Black (1) | Asian (2) | Indian (3) | Other (4) |
| **Glasses** | No Glasses (0) | Eye Wear (1) | Sunglasses (2) | | |
| **Beard** | Yes (0) | No (1) | | | |
| **Moustache** | Yes (0) | No (1) | | | |

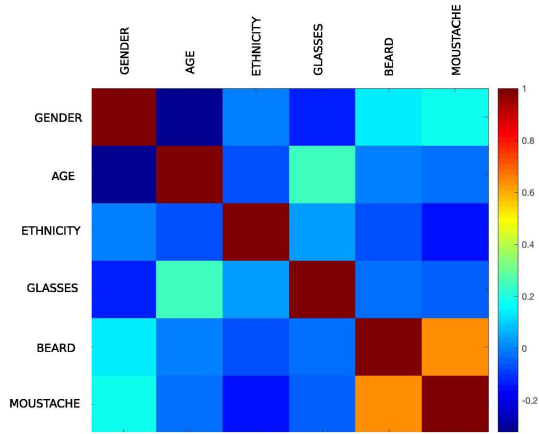

Fig. 2. **Correlation of soft biometrics**. The correlation is analyzed using the Pearson correlation coefficient.

*Other Mixture* otherwise. Regarding the age, we decided that a person belonged to a certain category based on the physical appearance, for instance, *Middle Aged* was asigned to anyone who looked in the range between $40 - 60$.

Based on the ground-truth, the statistics of LFW are as follows. Concerning gender, LFW contains 78% of *Male* and 22% of *Female* subjects. The majority of the population is *Middle Aged* 67%, followed by *Senior* 24% and *Youth* 13%. There is a small representation of *Baby* and *Child*, with less than 1% each. Regarding ethnicity, there is a superior percentage of *Caucasian* people (81.5%), with only 4% of them *Black*; 5.5% *Asian*; 2% *Indian* and 7% *Other* ethnicity. Most people show *No Glasses* (80%), while 18.6% have *Eyewear* and only 1.4% of them wear *Sunglasses*. With respect to facial hair, 6% and 10% of the LFW population has *Beard* or *Moustache*, respectively.

To gain insight about dependencies between soft biometrics from LFW, linear correlations between modalities are studied through the Pearson's linear correlation coefficient:

$$\text{corr\_coeff}(\mathbf{a}, \mathbf{b}) = \frac{\sigma_{\mathbf{ab}}}{\sigma_{\mathbf{a}}\sigma_{\mathbf{b}}} = \frac{E[(\mathbf{a} - \mu_{\mathbf{a}})(\mathbf{b} - \mu_{\mathbf{b}})]}{\sigma_{\mathbf{a}}\sigma_{\mathbf{b}}} \quad (1)$$

Correlations are depicted in Fig. 2. It is clearly visible that there is a high correlation between beard (yes 0; no 1) and moustache (yes 0; no 1), showing us that people with beard are very likely to have also moustache or vice versa. It is also logical to see some positive correlations between gender (male: 0; female: 1) and beard or moustache. There is also a moderate correlation between age (0: Baby; 1: Child; 2: Youth; 3: Middle Aged and 4 Senior) and glasses (0: No Glasses; 1: Eye Wear; 2: Sunglasses), giving us the idea that elder people tend to wear more eyewear or sunglasses, which is somewhat intuitive if we think that people lose their vision while growing. Finally there is moderate negative correlation between gender (Male: 0 ; Female: 1) and age (0: Baby; 1: Child; 2: Youth; 3: Middle Aged and 4 Senior), showing us that there are more older male than females.

### B. Automatic Estimation

We consider Face++[2] and Microsoft Cognitive Toolkit[3] Commercial Off-The-Shelf systems (COTS) to estimate soft biometrics. Face++ is a commercial face recognition system, which has achieved striking performance rates in the LFW face recognition competition (second best rate in the *unrestricted with labelled outside data* protocol with $0.9950 \pm 0.0036$ of mean accuracy). Apart from face recognition, both systems also provide common soft biometric estimation based on deep learning architectures. Microsoft Cognitive Toolkit is a deep learning framework developed by Microsoft Research that provides useful information based on vision, speech, language. No more information is available regarding these two COTS. The soft biometrics estimated by these systems are the following:

- **Gender**: *Male*, *Female* and confidence (Face++ & Microsoft)
- **Age:** value in years (Face++ & Microsoft)
- **Ethnicity:** *Caucasian*, *Black* and *Asian* (Face++)
- **Glasses:** *No Glasses*, *Eye Wear* and *Sunglasses* (Face++ & Microsoft)
- **Beard:** yes, no (Microsoft)
- **Moustache:** yes, no (Microsoft)

While age, gender and glasses are estimated by both COTS, ethnicity can only be estimated with Face++ and moustache and beard only with Microsoft Cognitive. Besides, age is automatically predicted as a continuous value in terms of number of years. As the groundtruth of age is described in categories, estimated values are thresholded in order to match continuous values to categories. Table IV reports the performance of automatic extraction algorithms of soft biometrics using Face++ and Microsoft Cognitive COTS. A superior performance is observed with the Microsoft Cognitive COTS for gender (92.94%) and age (59.25%). Glasses are better estimated with Face++ with an overall performance of 92.18%.

---
[2]We use the Official Matlab SDK For Face++ v2, which is the previous version to the one now available at https://www.faceplusplus.com/
[3]https://www.microsoft.com/cognitive-services/

TABLE IV
AUTOMATIC EXTRACTION OF SOFT BIOMETRICS. PERFORMANCE IS REPORTED IN TERMS OF ACCURACY FOR FACE++ AND MICROSOFT COGNITIVE COTS WITH RESPECT TO THE GROUNDTRUTH OBTAINED MANUALLY. N/A = NOT AVAILABLE

| Soft Biometric Trait - Instance | Face++ | Microsoft |
|---|---|---|
| Gender - Male | 92.15 | 93.48 |
| Gender - Female | 87.45 | 91.08 |
| Gender - Overall | 91.09 | 92.94 |
| Age - Baby | 100 | 100 |
| Age - Child | 53.22 | 45.16 |
| Age - Youth | 81.44 | 92.15 |
| Age - Middle Aged | 31.95 | 52.45 |
| Age - Senior | 33.42 | 59.62 |
| Age - Overall | 38.83 | 59.25 |
| Glasses - No Glasses | 93.46 | 94.47 |
| Glasses - Eye Wear | 89.62 | 82.19 |
| Glasses - Sunglasses | 55.43 | 65.22 |
| Glasses - Overall | 92.18 | 91.73 |
| Ethnicity - White | 88.26 | N/A |
| Ethnicity - Black | 76.24 | N/A |
| Ethnicity - Asian | 83.08 | N/A |
| Ethnicity - Overall | 87.44 | N/A |
| Beard - No | N/A | 94.39 |
| Beard - Yes | N/A | 87.28 |
| Beard - Overall | N/A | 93.97 |
| Moustache - Yes | N/A | 94.51 |
| Moustache - No | N/A | 90.42 |
| Moustache - Overall | N/A | 94.10 |
| Face Detection Rate | 98.29 | 94.22 |

TABLE V
EVALUATION OF AGE ACCURACY IN TERMS OF THE STANDARD DEVIATION ACHIEVED FOR SUBJECTS WITH MORE THAN ONE IMAGE IN THE LFW DATABASE. THE MEAN STANDARD DEVIATION IS THE AVERAGE OF THE STANDARD DEVIATION OBTAINED FOR ALL IDENTITIES WITH THE SAME NUMBER OF IMAGES

| # images/identity | # identities | Mean Standard Deviation | |
|---|---|---|---|
| | | Face++ | MS. |
| 1 | 4069 | 0 | 0 |
| 2 | 779 | 5.63 | 3.10 |
| 3 | 291 | 6.69 | 3.48 |
| 4 | 187 | 6.65 | 3.67 |
| 5 | 112 | 7.55 | 3.71 |
| 6 | 55 | 6.61 | 3.57 |
| 7 | 39 | 7.61 | 3.78 |
| 8 | 33 | 7.20 | 4.20 |
| 9 | 26 | 7.38 | 4.63 |
| 10 | 15 | 7.96 | 3.39 |
| 11 | 16 | 7.51 | 3.28 |
| 12 | 10 | 6.82 | 3.69 |
| 13 | 11 | 8.23 | 4.17 |
| 14 | 10 | 8.15 | 4.52 |
| 15 | 11 | 8.49 | 3.56 |
| more than 3 | 901 | 7.10 | 3.70 |

Ethnicity information is only available with Face++, and the possible instances that the COTS gives as outputs are: *Caucasian*, *Black* and *Asian*. Taking only into account images with groundtruth of these instances, the overall performance of ethnicity is around 87%. Likewise, the presence of beard and moustache can be only estimated through Microsoft Cognitive COTS, with an overall performance of 93.97% and 94.10%, respectively. Table IV also specifies the false detection rate obtained with both COTS. Face++ detects almost all faces while that Microsoft fails to detect more than 5% of them.

Table V describes an alternative way of reporting the performance attained by Microsoft and Face++ while predicting age. As both COTS predict age in terms of number of years, it is also possible to judge their reliability in terms of the standard deviation across estimated ages for a given identity. To this aim, we search the identities in the LFW database that contains a certain number of images. For instance, there are 4079 identities in the database that only have 1 image, 779 that contain 2 images, and so forth until 15 images. We do not consider the rest of the cases as very few individuals contain more than 15 images. Given a particular number of images per identity $x$, we compute the standard deviation obtained from the age estimation for all $x$ images of each identity. Later, the mean of all standard deviations attained for each individual is computed. Comparing this value for different number of images per identity can be seen that the standard deviation increases when more images/identity are considered. It is noticeable that Microsoft estimates better age than Face++. Concretely, when considering all identities with more than 3 images (roughly 901 identities), the standard deviation obtained with Face++ (7.10) is almost the double the standard deviation attained with Microsoft (3.70).

Note that the performances reported here are similar to some state-of-the-art results reported using the LFW+.[4] For example, in [16] they reported mean average error of 7.8% for age and 94% and 90% of accuracy for gender and ethnicity, respectively. Compared with the accuracies obtained in [22], it is observed that Face++ and/or Microsoft predicts gender, moustache and beard better than their approach (with 85.8%, 92.5% and 88.7% of accuracy, respectively) while age, ethnicity or glasses are better predicted with the approach by [22] (Age overall accuracy: 87.55%; Ethnicity overall accuracy: 92.95%; Glasses overall accuracy: 94.06%; ). It must be noted that the approach followed by Kumar *et al.* is based on binary classifiers, for instance regarding age there are 4 binary classifiers (child, youth, middle-aged, and senior).

## V. BIOMETRIC SYSTEMS

In this section we first overview the details concerning the soft biometric verification system. Later, the two face verification systems considered in this work as hard biometric systems are introduced.

### A. Soft Biometric Verification

This verification system uses soft biometrics as the only discriminative features. As depicted in Fig. 3, the soft biometric feature vector is a $N$-vector, where $N$ is the number of soft biometric modalities involved (between $1 - 6$). The particular value of each soft biometric trait is chosen according to the category labels shown in Table III. While gender, beard and moustache take binary values, glasses is divided into three categories and age and ethnicity into five different categories. Once the feature vector is built, dissimilarity is computed individually per each type of soft biometric trait through the Hamming distance (HD) for all soft biometrics except for the age, in which the Euclidean distance (ED) is more appropriate. HD just considers if the two labels are different or not, assigning 1 if they do not coincide, and 0 if they are equal. ED on the other hand measures how different two labels are since instances of age are ordered from younger to elder people. Knowing how much different two certain values are is more useful for recognition than just the fact that they are different; therefore ED is used. The overall distance of the whole set of soft biometrics is computed as the sum of all individual distances normalized by the particular number of soft biometrics. Finally, scores are computed by taking the inverse of the final dissimilarity.

---
[4]An extension of the LFW database that incorporates child images to have a more age-balanced dataset.

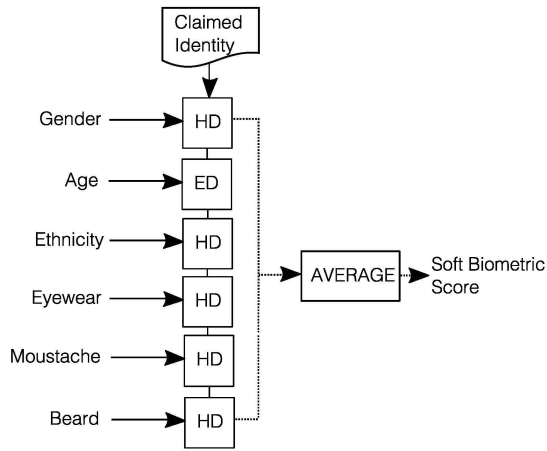

Fig. 3. **Soft Biometric Verification**. The soft biometric score is computed as the sum of the individual distances for each soft biometric.

Given that the proposed algorithm is based on various distances, the approach is quite robust against mislabeling. In particular, the only feature that may originate a small variability in the labeling process is Age (see Table III). Being that feature ordinal [37] makes the most probable mislabeling to occur between adjacent age groups, and therefore the effect of mislabeling would be very small after averaging distances as shown in Fig. 3. The remaining features are all nominal [37] and straightforward to label.

### B. Face Verification System

**Face++** is based on a structure of deep networks called Pyramid CNN [35]. This approach adopts a greedy-filter-and-down-sample operation enabling the training procedure to be very fast and computation efficient. The structure of the Pyramid CNN can naturally incorporate feature sharing across multi-scale face representations, increasing the discriminative ability of the resulting representation. Face++ COTS is used in order to obtain verification results. First, the faces to be compared are individually extracted by the detection module of the COTS. Once faces are detected, a similarity score is obtained by comparing the detected images through the comparison module of the Face++ COTS. There is not a public description of the particular implementation used by this COTS, so it is possible that it is not exactly the same as the system from which they report experiments in the competition.

The **VGG-face network** [29] was inspired by the previous VGG-Very-Deep-16 CNN network. It has been trained using a dataset of 2.6 million faces and 2622 classes (people). Their deep architecture comprises 39 layers from which 16 of them are conv-relu layers. When using a pre-trained model, there are two possible options: fine tune the pretrained model for a new application using a new target database, or use the pre-trained model as a feature extractor. In our case, as our aim is just to study the impact of soft and hard biometric fusion, we use the pre-trained model as a feature extractor. To this aim, first we need to resize images to the input size of the network, which is $224 \times 224$. The features are obtained by feedforwarding the LFW images until the $fc6$ layer, which turn out to be the layer that achieved better verification results (compared with the $fc7$ layer). This way, for each image we obtain a feature vector of 4096 elements. The matcher chosen in this case has been the cosine similarity ($score = cosine\_similarity(a, b)$, being $a$ and $b$ the feature vectors to be compared).

## VI. EXPERIMENTAL RESULTS

In this section, we study soft biometrics under different frameworks. First, the use of soft biometrics as *bag of soft biometrics* standalone for recognition is analyzed and empirical insights regarding the discrimination capability are given following the procedure described en Section V-A. Later, the use of *soft biometrics in conjunction with hard biometric systems* is addressed using both manual and estimated soft biometrics.

### A. Discrimination Analysis of Soft Biometrics

Among the different feature selection algorithms, the one employed in this work is the Sequential Floating Forward Selection (SFFS) algorithm, minimizing the EER [17]. This suboptimal searching technique is an iterative process. At each iteration, we build a soft biometric verification system as explained in Section V-A, but varying the particular set of features used (this choice is based on the results of previous subsets). This is done until the criteria does not improve. In our case the criteria is related to the performance of the system, in particular, to minimize the value of the EER. At each iteration, the SFFS algorithm is able to provide the most discriminative set of soft biometrics with a dimension specified by the user or with the dimension that gives the best value.

Given a particular target number of soft biometrics $N$ we employ the SFFS algorithm to find the set of $N$ soft biometrics (with $1 \leq N \leq 6$) that reaches jointly the best verification results. The concrete set of the $N$ most discriminative soft biometrics is included in Table VI. For instance, the most discriminative soft biometric trait is the age; age and ethnicity are the set of 2 soft biometric traits that minimizes the EER, and so forth. Table VI also shows the EER achieved for the different set of $N$ most discriminative soft biometrics $(SB_1, \ldots, SB_n)$, for both development and test set. As the number of soft biometrics considered increases, the verification performance also improves. The optimal set is achieved with the entire set of soft biometrics or with all but the beard. Another work in the literature achieved 6.3% EER when using 24 comparative attributes [5]. In our approach, we obtain 11.8% EER but a set of just 5 soft biometrics. Hence, we have empirically proved that person recognition based on soft biometrics may be possible under certain conditions (see Section II-A).

TABLE VI
VERIFICATION RESULTS USING A BAG OF SOFT BIOMETRICS. THE VERIFICATION SYSTEM IS BASED ON A SET OF THE $N \leq 6$ MOST DISCRIMINATIVE SOFT BIOMETRICS OBTAINED WITH SFFS FEATURE SELECTION. VERIFICATION RESULTS ARE REPORTED FOR THE DEVELOPMENT AND THE TEST SET OF $LFW$, IN TERMS OF $EER$ (%)

| N | Set of Soft Biometrics Considered | | | | | | Performance in terms of EER | |
|---|---|---|---|---|---|---|---|---|
| | $SB_1$ | $SB_2$ | $SB_3$ | $SB_4$ | $SB_5$ | $SB_6$ | Development | Test |
| 1 | Age | | | | | | 45.0 | 50.6 ± 3.1 |
| 2 | Age | Ethnicity | | | | | 22.0 | 31.1 ± 3.9 |
| 3 | Age | Ethnicity | Gender | | | | 14.0 | 19.1 ± 3.3 |
| 4 | Age | Ethnicity | Gender | Moustache | | | 10.0 | 14.4 ± 2.6 |
| 5 | Age | Ethnicity | Gender | Moustache | Glasses | | 8.0 | 11.8 ± 2.2 |
| 6 | Age | Ethnicity | Gender | Moustache | Glasses | Beard | 8.0 | 11.9 ± 2.2 |

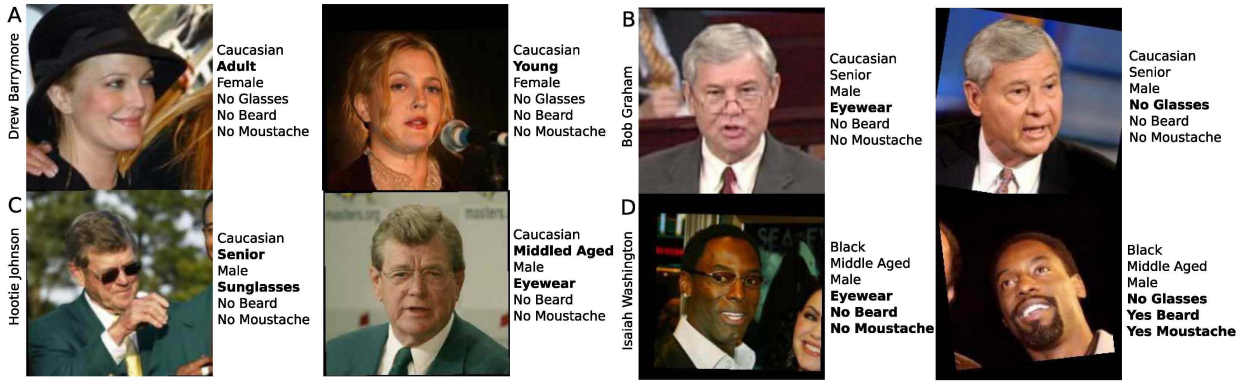

Fig. 4. **Verification examples based on a bag of soft biometrics.** The manual set of soft biometrics considered are gender, age, ethnicity, glasses, beard and moustache. Visual examples regarding the challenges regarding reliability of age, intra-class variations of glasses, beard and moustache are shown.

Even if we observe clear benefits from using soft biometrics for person recognition, there are still some issues. For example, we have observed in the manual labelling of soft biometrics some differences between age estimation of different images from the same subject. The predominant confusion occurs between *Middle Aged* and *Senior* instances. While it is true that it can be due to mistakes from the annotator, the image appearance, facial expression and other factors also play an important role (see the case of *Drew Barrymore* in Fig. 4.A). Concerning the use of glasses, even if the best results are achieved when glasses are included, there are some things to note. The different instances of glasses were *No Glasses*, *Eye Wear* or *Sunglasses*. First, we have observed that information regarding the use of sunglasses might not be used as soft biometrics unless someone wear sunglasses all the time, which is not the normal case (see *Hootie Johnson* example in Fig. 4.C). Second, information regarding eyewear glasses may be potentially useful if someone always wear them. It is also very common to find people wearing equally eyewear glasses or contact lenses or people wearing only eyewear glasses while reading (see *Bob Graham* in Fig. 4.B). The same way, facial hair information might be useful for recognition if it remains throughout time. In some cases, facial attributes such as beard and moustache are not completely reliable, since they are attributes that may change often. This explains why moustache, glasses and beard are ranked after gender by SFFS. While gender information tends to be almost invariant, these soft biometrics may change across time. Depending on the application, such intra-class variations can have an impact on the performance or not, e.g. in people re-identification applications the time span is so small that it is less likely that these soft biometrics change.

### B. Combining Hard and Soft Biometric Systems

In this section we aim to study the convenience of fusing hard and soft information to enhance the verification task. We use the two state-of-the-art face verification systems described in Section V-B and combine them with soft information extracted manually or automatically. This way, we will investigate to what extent soft biometrics complement the evidence given by hard biometric systems. In this case, both modalities are fused at score level with equal weights.

*1) Results With Manual Soft Biometrics:* Table VII reports the performance of the fusion between soft and hard biometric systems when using manual soft biometrics. We study this fusion approach with the selected set of $N$ soft biometrics provided by the SFFS algorithm in Section VI-A, with $N = 1 \ldots 6$. For comparison purposes, individual verification results from the different configurations of soft biometrics and the two considered face-based verification are also included. From Table VII we can see that the best face verification system is VGG-face. It can be also observed that the inclusion of soft biometrics enhance the face system in all cases, even when considering just one soft biometric trait. The best results are obtained when considering the set of age, ethnicity, gender, moustache and glasses. The introduction of soft biometric information provides up to 40% of relative improvement for both face verification systems, which is very remarkable. As already discussed in Section VI-A, the use of *Sunglasses* as a possible instance of glasses is not very appropriate. Last two rows from Table VII report the results attained when *Sunglasses* are discarded, reaching additional improvements with respect to their counterpart configurations in which *Sunglasses* was included.

Fig. 5 depicts some visual examples of cases in which the proposed fusion approach is able to improve the hard biometric system (success cases) and others in which the fusion approach still fails (failure cases). Success and failure cases are studied under the point of view of genuine and impostor comparisons. Success cases with respect to genuine comparisons happen when genuine comparisons are wrongly detected as impostor by the face verification system but correctly predicted as genuine by the fusion system. In those cases, soft biometrics yield to correctly matched genuine pairs when matching through face is hindered by illumination and pose (see *Gillian Anderson* in Fig. 5.A) or changes in appearance and expression (see *Sandra Bullock* in Fig. 5.C). Conversely, failure cases with respect to genuine comparisons occur when those are wrongly predicted as impostor by both face verification and fusion systems (see *Juan Pablo Montoya* in Fig. 5.D or *Angela Basset* Fig. 5.F). In those cases, the face score is far from the threshold (due to severe changes between the compared images) and therefore soft biometrics are not enough to change the decision. Additionally, the mismatch between soft biometric labels prevents the overall system from detecting correctly positive matches, especially for scores close to the threshold (*Juan Pablo Montoya*). It is also worth noting that there can be cases in which genuine comparisons are predicted as genuine by the face verification system but wrongly predicted by the fusion system. These sort of cases are a minority.

Fig. 5 also depicts success and failure cases concerning impostor comparisons. In this regard, success cases comprise impostor comparisons wrongly detected as genuine by the face system but correctly predicted as impostor by the fusion system (see for instance *Dianne Reeves-Larry Willmore* in Fig. 5.H). In these cases the face system thinks that two face images belong to the same identity due to similar appearance, expression or context. By incorporating soft biometric information that does not match between the individuals, these comparisons successfully turn into impostor. On the other hand, failure cases regarding impostor comparisons happen when those are wrongly predicted as genuine comparisons by both face verification and fusion systems. In this case, the face system also mismatches identities caused by resembling appearances. However, soft biometrics are unable to help as a consequence that the two different individuals share fully or partially the same soft biometric profile (see for instance *Gian Marco-Kevin Stallings* in Fig. 5.J). More insight regarding this cross-subject interference can be found in [8].

*2) Results With Automated Soft Biometrics:* The following step in our study is to analyze the fusion of soft and hard biometrics when soft biometrics are automatically extracted and hence it is possible to encounter soft biometrics incorrectly estimated. We previously decided which COTS to use for each soft biometric trait. This decision is made based on the COTS with better performance. According to the accuracies reported in Table IV, gender, age, moustache and beard are estimated by the Microsoft system while ethnicity and glasses are estimated using Face++. Unlike the case using manual soft biometrics in which age information was addressed as a categorical variable in 5 age groups, here age information is used in years. Results are reported in Table VIII. First of all we notice that the best performance for a bag of soft biometrics degrades to around 20% EER when using the automatic estimations, which is reasonable taking into account that the work conducted in [22] reached 14.25% EER while using 73 attributes automatically estimated.

Table VIII also shows the results achieved when fusing face information and the set of estimated soft biometrics. In this case, the fusion approach outperforms the face verification system when considering more than 2 soft biometric traits (Face++) or 3 soft biometric traits (VGG-face). The best improvements are achieved when considering age, gender, ethnicity, moustache and glasses (without sunglasses). Concretely, the performance of face verification system is reduced from 12.7% to 11.4% and from 7.8% to 6.6% when considering also soft biometrics for Face++ and VGG-face, respectively. The relative improvement of the fusion with respect to the face verification system standalone is of 10.23% for Face++ and 15.38% for VGG-face, respectively. Table IX reports the Mean Classification Accuracies achieved for the best configuration of soft biometrics (age, gender, ethnicity, moustache and glasses*) using the set of manual and automatic labels and both face recognition systems. Comparing our results with those reported by Zhang *et al.* [41], we would like to note some differences and similarities. First, datasets employed in both studies are different in terms of number of subjects (LFW has a larger number of subjects), and in terms of acquisition conditions (LFW is more unconstrained than GBU). Secondly, gender and race are used both in [41] and here in our work. While we have carried out a previous study to understand the discrimination capabilites of our set of soft biometrics (figuring out that age was more discriminative, followed by ethnicity and gender), the work done by Zhang *et al.* do not explore the discrimination capabilites of their particular set of soft biometrics.

Even if our results empirically prove that hard biometric systems can benefit from the incorporation of soft biometrics estimated automatically, there is considerable room for improvement. We observe a large difference in the relative improvements of face verification when introducing manual (ca. 40%) or automatic soft biometrics (ca. 15%). This can be explained first of all because of the limited accuracy in estimating the soft biometrics by the COTS systems. The accuracies achieved for each soft biometric trait are: 93.01% for gender; 92.18% for glasses; 87.44% for ethnicity; 93.97% for beard and 94.10% for moustache. Age is estimated by

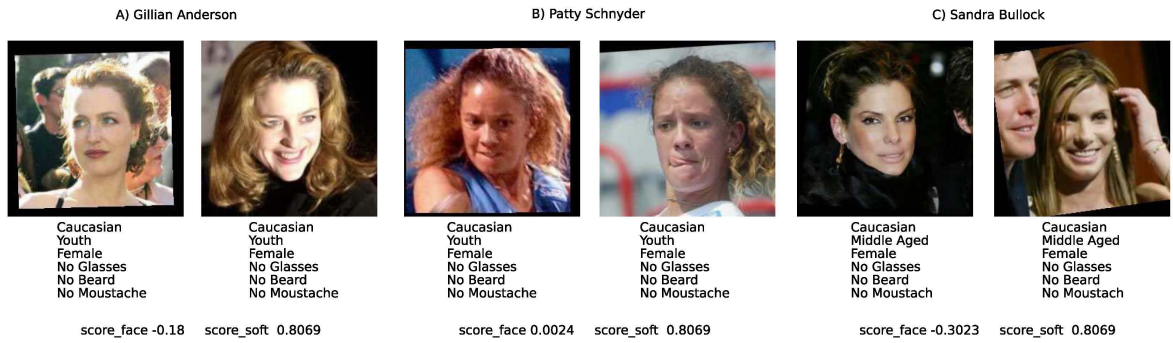
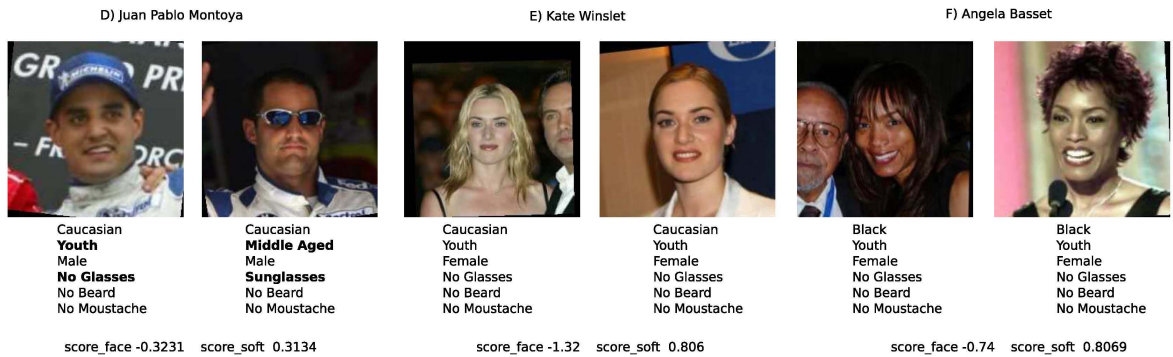
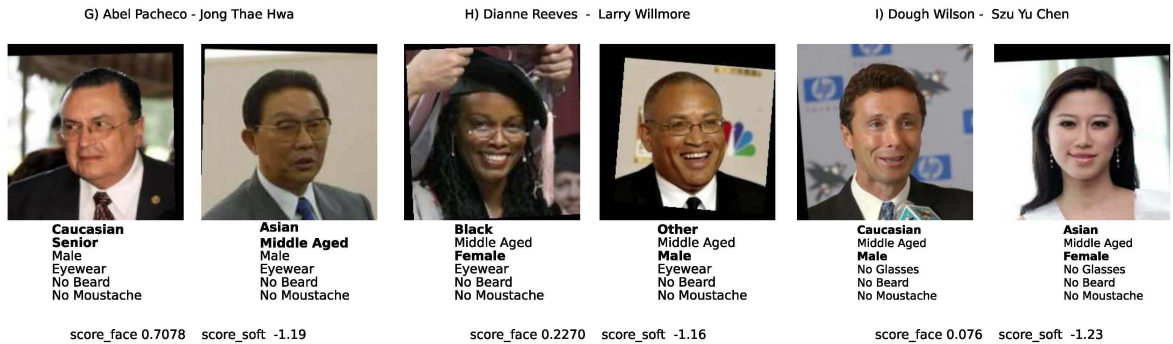
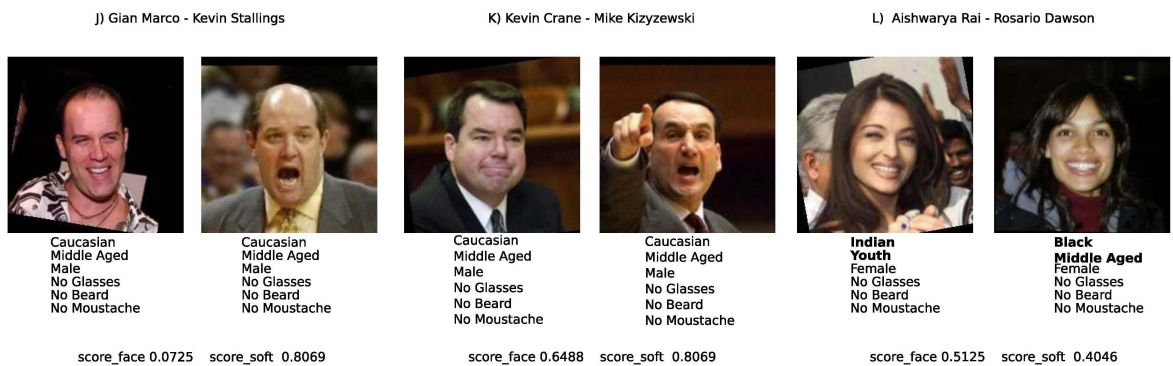

Fig. 5. **Qualitative results of the fusion between manual soft biometrics and face**. Examples are divided into success cases and failure cases of genuine and impostor comparisons. The face verification system considered is the VGG-face, with a threshold of 0.0067. The set of soft biometrics considered is ethnicity, age, gender, glasses, beard and moustache.

Microsoft Cognitive COTS with a mean standard deviation of 3.70. Second, the number of possible ethnicity instances estimated with Face++ is reduced to 3 as opposed to the 5 different instances considered in the manual approach. Fusion results can be further improved by considering a larger number of ethnicity instances. There results motivate further

TABLE VII

FUSION OF MANUAL SOFT BIOMETRICS WITH HARD BIOMETRICS. TWO HARD BIOMETRIC SYSTEMS ARE CONSIDERED: A FACE VERIFICATION SYSTEM BASED ON FACE++ COTS AND A FACE VERIFICATION SYSTEM BASED ON VGG-FACE DEEP LEARNING. SOFT BIOMETRICS HAVE BEEN MANUALLY EXTRACTED. ALL VERIFICATION RESULTS ARE REPORTED FOR THE TEST SET OF LFW, IN TERMS OF EQUAL ERROR RATE (EER IN %) AND USING THE EXPERIMENTAL PROTOCOL DEFINED IN SECTION III. * MEANS THAT THE SUNGLASSES HAVE BEEN DISCARDED AS INSTANCE FROM GLASSES SOFT BIOMETRIC

| Performance of Soft Biometrics | | Face | | Fusion | |
|---|---|---|---|---|---|
| Set of Soft Biometrics | | Face++ | VGG-face | Face++ | VGG-face |
| Age | 50.6 ± 3.1 | 12.7 ± 1.4 | 7.8 ± 1.2 | 10.9 ± 1.4 | 7.1 ± 0.7 |
| Age Ethnicity | 31.1 ± 3.9 | | | 9.0 ± 1.2 | 5.8 ± 0.5 |
| Age Ethnicity Gender | 19.1 ± 3.3 | | | 8.4 ± 1.3 | 4.9 ± 0.6 |
| Age Ethnicity Gender Moustache | 14.4 ± 2.6 | | | 7.7 ± 1.5 | 4.8 ± 0.5 |
| Age Ethnicity Gender Moustache Glasses | 11.9 ± 2.2 | | | 7.7 ± 1.5 | 4.8 ± 0.7 |
| Age Ethnicity Gender Moustache Glasses Beard | 12.0 ± 2.2 | | | 8.3 ± 1.7 | 5.4 ± 0.9 |
| Age Ethnicity Gender Moustache Glasses* | 11.2 ± 2.1 | | | 7.6 ± 1.4 | 4.4 ± 0.5 |
| Age Ethnicity Gender Moustache Glasses* Beard | 11.1 ± 2.1 | | | 8.0 ± 1.7 | 5.2 ± 0.7 |

TABLE VIII

FUSION OF ESTIMATED SOFT BIOMETRICS WITH HARD BIOMETRICS. TWO HARD BIOMETRIC SYSTEMS ARE CONSIDERED: A FACE VERIFICATION SYSTEM BASED ON FACE++ COTS AND A FACE VERIFICATION SYSTEM BASED ON THE VGG-FACE DEEP LEARNING. SOFT BIOMETRICS HAVE BEEN AUTOMATICALLY EXTRACTED FROM FACE++ AND MICROSOFT COTS. ALL VERIFICATION RESULTS ARE REPORTED FOR THE TEST SET OF LFW, IN TERMS OF EQUAL ERROR RATE (EER IN %) AND USING THE EXPERIMENTAL PROTOCOL DEFINED IN SECTION III. * MEANS THAT THE SUNGLASSES HAVE BEEN DISCARDED AS INSTANCE FROM GLASSES SOFT BIOMETRIC

| Performance of Soft Biometrics | | Face | | Fusion | |
|---|---|---|---|---|---|
| Set of Soft Biometrics | | Face++ | VGG-face | Face++ | VGG-face |
| Age | 27.2 ± 1.6 | 12.7 ± 1.4 | 7.8 ± 1.2 | 13.7 ± 1.7 | 10.3 ± 1.2 |
| Age Ethnicity | 25.8 ± 2.5 | | | 12.9 ± 1.6 | 8.8 ± 0.7 |
| Age Ethnicity Gender | 22.2 ± 1.8 | | | 11.9 ± 1.9 | 8.1 ± 0.6 |
| Age Ethnicity Gender Moustache | 21.6 ± 2.0 | | | 11.7 ± 1.9 | 7.3 ± 0.8 |
| Age Ethnicity Gender Moustache Glasses | 22.6 ± 2.1 | | | 11.6 ± 1.9 | 6.8 ± 0.7 |
| Age Ethnicity Gender Moustache Glasses Beard | 23.8 ± 1.9 | | | 11.8 ± 1.8 | 6.9 ± 1.0 |
| Age Ethnicity Gender Moustache Glasses* | 22.7 ± 1.9 | | | 11.4 ± 1.9 | 6.6 ± 0.8 |
| Age Ethnicity Gender Moustache Glasses* Beard | 24.1 ± 1.7 | | | 11.5 ± 1.7 | 6.8 ± 1.0 |

TABLE IX

MEAN CLASSIFICATION ACCURACIES (IN %)

| | Soft | | Fusion | |
|---|---|---|---|---|
| Hard | Manual | Automatic | Manual | Automatic |
| Face++: 83.95 | 88.83 | 77.23 | 91.83 | 88.66 |
| VGG-face: 92.20 | | | 94.83 | 93.30 |

research toward more accurate soft biometric automatic extraction algorithms that can complement hard-based biometric systems.

## VII. FINAL DISCUSSION AND CONCLUSION

In this work soft biometrics have been studied for person recognition in unconstrained scenarios. To this end, we have manually labeled all images from the LFW database in terms of gender, age, ethnicity, glasses, beard and moustache.

Our first aim was to investigate the discrimination capability of each soft biometric trait. By employing the SFFS algorithm we have learned that soft biometrics are ranked from the most discriminative to the least discriminative in the following order: age, ethnicity, gender, moustache, glasses, beard. Soft biometrics such as moustache, glasses and beard are less discriminative due to their intra-class variations.

In order to assess to what extent soft biometrics can complement hard biometric modalities, this rank of soft biometrics was further combined with two state-of-the-art face recognition systems (COTS Face++ and VGG-face). We first employ the manual estimation of soft biometric modalities. In all the different configurations of face verification systems and the set of the $N$ most discriminative soft biometrics ($N$ from 1 to 6), improved results are always achieved with respect to the hard biometric system standalone. The best fusion results are obtained when considering the entire set of soft biometrics, reaching relative performance improvements over the face systems of up to 40%.

Then, the fusion of the face verification systems is analyzed along with a set of automatically estimated soft biometrics. We empirically prove that estimated soft biometrics improve face verification systems although the relative performance improvement in this case is reduced to $10 - 15\%$. According to Table II, this is one of the first works that have studied the potential of automatic soft biometrics for person recognition in fully unconstrained scenarios using state-of-the-art face verification systems, concluding that considerable improvements can be gained if soft biometrics are estimated accurately. In this regard, there is still a large room for improvement in the area of automatic algorithms before fully exploiting the potential of soft biometrics for scenarios in-the-wild.

Besides, we foresee additional improvements by weighting the different soft biometrics according to their discrimination capability, permanence level and/or accuracy estima-tion [7] in a quality-based fusion approach [6]. Additionally, alternative fusion schemes between hard and soft biometric systems are noteworthy for further exploration and study [11]. We also believe that additional improvements can be gained by using more sophisticated deep learning architectures to estimate soft biometrics [15]. For instance, age estimation could be improved by a multi-task CNN architecture that jointly estimates gender and age. Another limitation of the annotated dataset made publicly available[5] is the use of a single annotator. It may be interesting additional targeted research towards understanding the factors and implications when having diverse annotators, perhaps through crowd-sourcing [2].

[5]https://atvs.ii.uam.es/atvs/LFW_SoftBiometrics.html

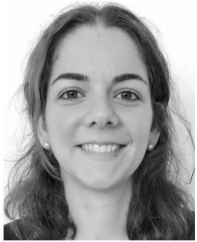

**Ester Gonzalez-Sosa** received the B.S. degree in computer science and the M.Sc. degree in electrical engineering from the Universidad de Las Palmas de Gran Canaria in 2012 and 2014, respectively, and the Ph.D. degree from the Biometric Recognition Group, Universidad Autonoma de Madrid, in 2017. In 2017, she joined the Distributed Reality Solutions Laboratory, Nokia Bell-Labs. She has carried out several research internships in worldwide leading groups in biometric recognition, such as TNO, EURECOM, or Rutgers University. Her research interests include pattern recognition, signal processing, and biometrics, with emphasis on face, body, soft biometrics, and millimeter imaging. She has been the recipient of the UNITECO AWARD from the Spanish Association of Electrical Engineers and the competitive Obra Social La CAIXA Scholarship.

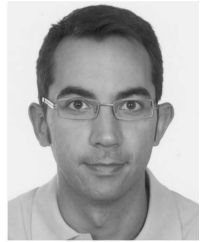

**Ruben Vera-Rodriguez** received the M.Sc. degree in telecommunications engineering from the Universidad de Sevilla, Spain, in 2006, and the Ph.D. degree in electrical and electronic engineering from Swansea University, U.K., in 2010. Since 2010, he has been with the Biometric Recognition Group, Universidad Autonoma de Madrid, Spain, first as a recipient of the Juan de la Cierva Post-Doctoral Fellowship from the Spanish Ministry of Innovation and Sciences, and then as an Assistant Professor since 2013. His research interests include signal and image processing, pattern recognition, and biometrics, with emphasis on signature, face and gait verification, and forensic applications of biometrics. He is actively involved in several national and European projects focused on biometrics.

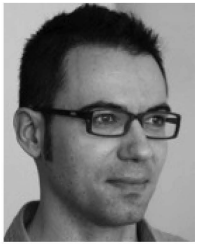

**Julian Fierrez** received the M.Sc. and Ph.D. degrees in telecommunications engineering from the Universidad Politecnica de Madrid, Spain, in 2001 and 2006, respectively. Since 2002, he has been affiliated with the Biometric Recognition Group (ATVS), first at the Universidad Politecnica de Madrid, and since 2004, at Universidad Autonoma de Madrid, where he is currently an Associate Professor. From 2007 to 2009, he was a Visiting Researcher with Michigan State University, USA, under the Marie Curie Fellowship. His research interests include general signal and image processing, pattern recognition, and biometrics, with emphasis on signature and fingerprint verification, multi-biometrics, biometric databases, system security, and forensic applications of biometrics. He has been actively involved in multiple EU projects focused on biometrics (e.g., TABULA RASA and BEAT) and has attracted notable impact for his research. He was a recipient of a number of distinctions, including the EBF European Biometric Industry Award 2006, the EURASIP Best Ph.D. Award 2012, the Medal in the Young Researcher Awards 2015 by the Spanish Royal Academy of Engineering, and the Miguel Catalan Award to the Best Researcher under 40 in the Community of Madrid in the general area of Science and Technology. Since 2016, he has been an Associate Editor of the IEEE TRANSACTIONS ON IFS and the IEEE BIOMETRICS COUNCIL NEWSLETTER.

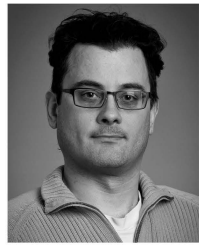

**Fernando Alonso-Fernandez** received the M.S. and Ph.D. degrees in telecommunications engineering from the Universidad Politecnica de Madrid, Spain, in 2003 and 2008, respectively. Since 2010, he has been with the Centre for Applied Intelligent Systems Research, Halmstad University, Sweden, first as a recipient of the Marie Curie IEF and the Post-Doctoral Fellowship from the Swedish Research Council and later as a recipient of a Project Research Grant for Junior Researchers of the Swedish Research Council. Since 2017, he has been an Associate Professor with Halmstad University. He has been actively involved in multiple EU (such as FP6 Biosecure NoE and COST IC1106) and national projects focused on biometrics and human–machine interaction. He has over 70 international contributions at refereed conferences and journals and has authored several book chapters. His research interests include signal and image processing, pattern recognition, and biometrics, with emphasis on facial cues and body biosignals. He Co-Chaired ICB2016 and the 9th IAPR International Conference on Biometrics.